\documentclass{article}

\PassOptionsToPackage{numbers, compress}{natbib}

\usepackage[preprint]{neurips_2026}

\usepackage[utf8]{inputenc} 
\usepackage[T1]{fontenc}    
\usepackage{hyperref}       
\usepackage{url}            
\usepackage{booktabs}       
\usepackage{amsfonts}       
\usepackage{nicefrac}       
\usepackage{microtype}      
\usepackage{xcolor}         

\usepackage{amsmath}
\usepackage{amssymb}
\usepackage{graphicx}
\usepackage{multirow}
\usepackage{makecell}

\title{Text-Guided Medical Image Segmentation via Residual Aggregation and Dense Alignment Representations}

\author{%
Jie Gui\textsuperscript{1,2},
Hang Tu\textsuperscript{2},
Yunfeng Dou\textsuperscript{3},
Wen Sha\textsuperscript{1,2}\thanks{Corresponding authors.},
Xiuquan Du\textsuperscript{1,3}\footnotemark[1],
Shuo Li\textsuperscript{4}
\\[1.2mm]
\textsuperscript{1}Key Laboratory of Intelligent Computing and Signal Processing,\\
Ministry of Education, Anhui University, Hefei, China\\
\textsuperscript{2}School of Artificial Intelligence, Anhui University, Hefei, China\\
\textsuperscript{3}School of Computer Science and Technology, Anhui University, Hefei, China\\
\textsuperscript{4}Department of Biomedical Engineering,
Case Western Reserve University, Cleveland, OH, USA\\[0.8mm]
\texttt{\{w125111032,w125121009,e125121027\}@stu.ahu.edu.cn}\\
\texttt{\{04069,dxqllp\}@ahu.edu.cn}
}

\begin{document}

\maketitle

\begin{abstract}
  Specifying segmentation targets with natural language provides flexible semantic conditioning for medical image segmentation. However, existing methods still struggle to reliably ground high-level textual semantics into accurate pixel-level spatial prompts, especially under low-contrast lesions, ambiguous boundaries, and visually similar anatomical contexts. In this paper, we propose \textbf{RADAR} (\textbf{R}esidual \textbf{A}ggregation and \textbf{D}ense \textbf{A}lignment \textbf{R}epresentations), a parameter-efficient framework for text-guided medical image segmentation. RADAR explicitly decouples and synergizes the medical semantic anchoring capability of BiomedCLIP with the prompt-driven mask decoding ability of MedSAM2, while enabling low-cost task adaptation via LoRA. Specifically, Guided Residual Depth Aggregation adaptively aggregates cross-layer features under semantic guidance to alleviate delayed textual injection and deep semantic dilution. Structured Dense Representation Alignment leverages a frozen BiomedCLIP to construct crop-level target, boundary, and context anchors, improving the local discriminability of dense features and the reliability of automatic prompts. Multi-modal Prompt Generation further converts cross-modal semantic responses into composite text, point, and mask prompts for MedSAM2 decoding. Experiments on three public medical image segmentation datasets show that RADAR achieves competitive performance with only 0.74M trainable parameters, demonstrating its effectiveness in parameter-efficient adaptation, semantic grounding, and dense representation enhancement. The source code is available at \url{https://github.com/JGuiidev/RADAR}.
\end{abstract}

\section{Introduction}

\begin{figure}[t]
  \centering
  \includegraphics[width=0.95\linewidth]{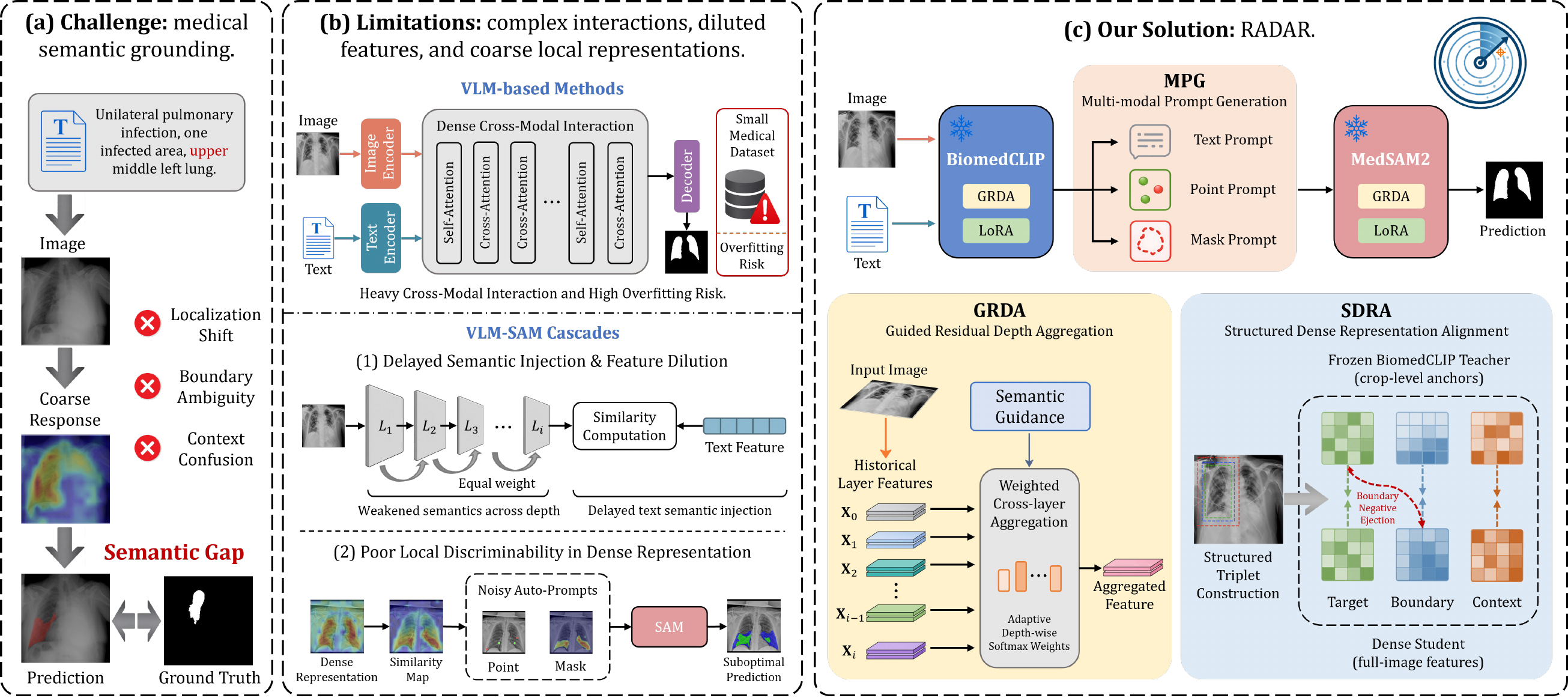}
  \caption{
  Challenge and limitations of existing methods, and our solution.
  }
  \label{fig:motivation}
\end{figure}

Text-guided medical image segmentation dynamically specifies targets via natural language. Unlike closed-set paradigms relying on predefined labels, this task better aligns with real-world clinical scenarios involving diverse pathologies and low-quality imaging~\citep{koleilat2026medclipseg,UAVLS2026}. However, natural language typically provides only high-level semantic priors, such as lesion categories, anatomical locations, and regional scopes, whereas segmentation requires precise pixel-level localization and boundary delineation. This discrepancy often leads to localization deviation, boundary ambiguity, and contextual confusion, forming a substantial semantic-to-spatial gap, as illustrated in Figure~\ref{fig:motivation}(a). Thus, reliably translating medical textual semantics into executable spatial prompts remains a key challenge.

Vision-Language Models (VLMs) provide a foundation for this semantic grounding. Early end-to-end multimodal medical segmentation methods rely on elaborate dense cross-modal interactions~\citep{WanQij2025towards,huang2025cross}, which often introduce high training costs and may overfit under limited medical annotations. Conversely, prompt-driven models such as SAM and MedSAM exhibit strong class-agnostic mask generation and can recover accurate boundaries from point or box prompts~\citep{kirillov2023segment,MedSAM2024}. Yet, they lack native text understanding and depend heavily on manually specified visual prompts.

Recently, foundation models such as CLIP~\citep{radford2021learning} and BiomedCLIP~\citep{zhang2023biomedclip} have leveraged contrastive learning to enhance cross-modal alignment, mapping textual and visual representations into a shared semantic space. Several cascaded architectures further combine VLMs with SAM, using VLMs to provide semantic spatial cues and SAM for fine-grained mask decoding~\citep{zhang2025evfsam}. Although these approaches exploit their complementary capabilities, existing VLM-SAM frameworks still face two critical bottlenecks, as summarized in Figure~\ref{fig:motivation}(b).

First, textual semantics participate too late, while historical visual evidence can be diluted during deep propagation. Dual-encoder VLMs typically encode visual and textual inputs independently and establish their correspondence only through deep similarity computation or late cross-modal interaction. Consequently, text cannot explicitly determine which intermediate visual representations should be retained for the current target during encoding. Meanwhile, conventional layer-wise feature accumulation often assigns fixed weights to historical representations, gradually weakening discriminative intermediate-layer responses during deep propagation~\citep{chen2026attnres}.

Second, VLM dense representations lack sufficient local discriminability. VLMs are primarily optimized for image-level or global semantic matching, and their local dense features are not explicitly trained for pixel-level target discrimination~\citep{wu2024clipself}. Under low-contrast lesions, blurred boundaries, and visually similar neighboring anatomical structures, text-relevant responses may simultaneously activate target, boundary, and contextual regions. Such ambiguity increases noise in automatically generated prompts and consequently limits the mask decoding quality of downstream SAM models.

To address these limitations, we propose RADAR (Residual Aggregation and Dense Alignment Representations), a parameter-efficient framework for text-guided medical image segmentation. As shown in Figure~\ref{fig:motivation}(c), RADAR decouples the medical vision-language semantic anchoring capability of BiomedCLIP from the prompt-conditioned mask decoding capability of MedSAM2, while introducing LoRA into key branches for parameter-efficient adaptation~\citep{hu2022lora}. Guided Residual Depth Aggregation (GRDA) uses textual semantics as a depth-routing condition to adaptively select and aggregate historical visual evidence at each spatial location, rather than repeatedly injecting textual content into the current visual representation. This text-conditioned depth routing mitigates delayed semantic interaction and feature dilution during deep propagation. Structured Dense Representation Alignment (SDRA) further improves the local discriminability of VLM dense representations by constructing target, boundary, and context anchors with a frozen BiomedCLIP teacher. Unlike generic global alignment~\citep{wu2024clipself}, SDRA emphasizes fine-grained regional differences and thereby improves the reliability of spatial responses. Finally, Multi-modal Prompt Generation (MPG) serves as the semantic-to-spatial interface between BiomedCLIP and MedSAM2, converting the enhanced cross-modal responses into composite text, point, and mask prompts for prompt-conditioned decoding.

The main contributions of this work are summarized as follows:

\begin{itemize}
    \item We propose RADAR, a parameter-efficient framework integrating BiomedCLIP for semantic anchoring and MedSAM2 for mask decoding. MPG seamlessly translates cross-modal responses into composite spatial and textual prompts.

    \item We introduce GRDA, a semantic-guided depth aggregation mechanism that mitigates delayed cross-modal interaction and semantic dilution in deep feature propagation.

    \item We propose SDRA, which constructs crop-level semantic anchors with a frozen BiomedCLIP to improve dense feature discriminability and prompt reliability.
\end{itemize}

\section{Related work}

\subsection{Vision-language models in medical segmentation}

Traditional medical image segmentation methods, such as FCN~\citep{long2015FCN}, U-Net~\citep{Ronneberger2015UNet}, and TransUNet~\citep{Chen2024TransUNet}, mainly rely on predefined label spaces and thus struggle with open-vocabulary targets and weakly standardized clinical descriptions. Methods such as LAVT~\citep{yang2025LAVT} and LViT~\citep{li2024LViT} incorporate language guidance through Vision Transformers~\citep{dosovitskiy2021an} and cross-modal attention. However, dense cross-modal interaction usually increases training costs and may overfit under limited medical annotations.

Contrastive VLMs provide an alternative foundation for medical semantic grounding. CLIP~\citep{radford2021learning} and ALIGN~\citep{jia2021scaling} learn shared image-text spaces, while MedCLIP~\citep{wang2022medclip} and BiomedCLIP~\citep{zhang2023biomedclip} further improve medical image-text alignment using domain-specific data. Nevertheless, their dense local features are not explicitly trained for pixel-level target discrimination~\citep{wu2024clipself,zhou2022extract}. Directly using VLM similarity responses for segmentation can therefore cause ambiguous activations, boundary leakage, and contextual false responses.

\subsection{Promptable foundation models and VLM-SAM cascades}

SAM~\citep{kirillov2023segment} and its medical extensions, including MedSAM~\citep{MedSAM2024} and MedSAM2~\citep{ma2025medsam2segment3dmedical}, provide strong prompt-driven mask generation but lack native text understanding. Recent methods therefore combine VLM-based semantic localization with SAM-based mask decoding. In natural images, EVF-SAM~\citep{zhang2025evfsam}, GLaMM~\citep{rasheed2024glamm}, and DIT~\citep{huang2024deep} explore language grounding or instruction tuning. In the medical domain, MedCLIP-SAMv2~\citep{KOLEILAT2025medclipsam}, TGSAM-2~\citep{YuaRun2025tgsam2}, SegMoTE~\citep{lu2026segmote}, and LGA~\citep{Hu_LGA_MICCAI2024} further explore text-guided SAM adaptation. Parameter-efficient techniques such as LoRA~\citep{hu2022lora}, QLoRA~\citep{dettmers2023qlora}, and MAdapter~\citep{Zha_MAdapter_MICCAI2024} reduce the cost of adapting these large models.

Existing approaches typically either fuse language repeatedly through dense cross-modal interactions or directly convert the final VLM response into SAM prompts. The former incurs substantial interaction costs, whereas the latter is limited by late semantic interaction and noisy dense responses. In particular, point or mask prompts directly derived from VLM responses~\citep{lee2025Effective} may contain background noise, motivating further improvement of semantic propagation and spatial discrimination before prompt generation.

\subsection{Feature aggregation and dense representation alignment}

\paragraph{Semantic aggregation in deep networks.}

Residual connections and PreNorm structures are widely used to stabilize deep networks~\citep{ba2016layer,he2016deepresidual}, but fixed residual accumulation can progressively dilute the contribution of individual layers. Attention Residuals~\citep{chen2026attnres} use depth-wise attention to aggregate historical representations, while BiPVL-Seg~\citep{sultan2025bipvlseg} progressively injects language information through vision-language fusion. Unlike such cross-modal feature fusion or modality-agnostic depth aggregation, GRDA uses text as a depth-routing condition to select historical visual evidence at each spatial location, enabling text-conditioned and spatially resolved cross-layer retention.

\paragraph{Region-level alignment for dense representations.}

VLM dense features often lack local discriminability when transferred to segmentation tasks. CLIPSelf~\citep{wu2024clipself} and CAT-Seg~\citep{cho2024cat} improve dense prediction through self-distillation and cost aggregation, respectively. Unlike generic alignment that mainly encourages teacher-student feature consistency, SDRA explicitly constructs target, boundary, and context regions and models their relative relationships. Its boundary-negative rejection prevents boundary representations from becoming as target-like as the target itself, thereby enforcing relative regional discrimination among target bodies, ambiguous boundaries, and adjacent contexts.

\section{Methodology}
 
\begin{figure}[t]
  \centering
  \includegraphics[width=0.95\linewidth]{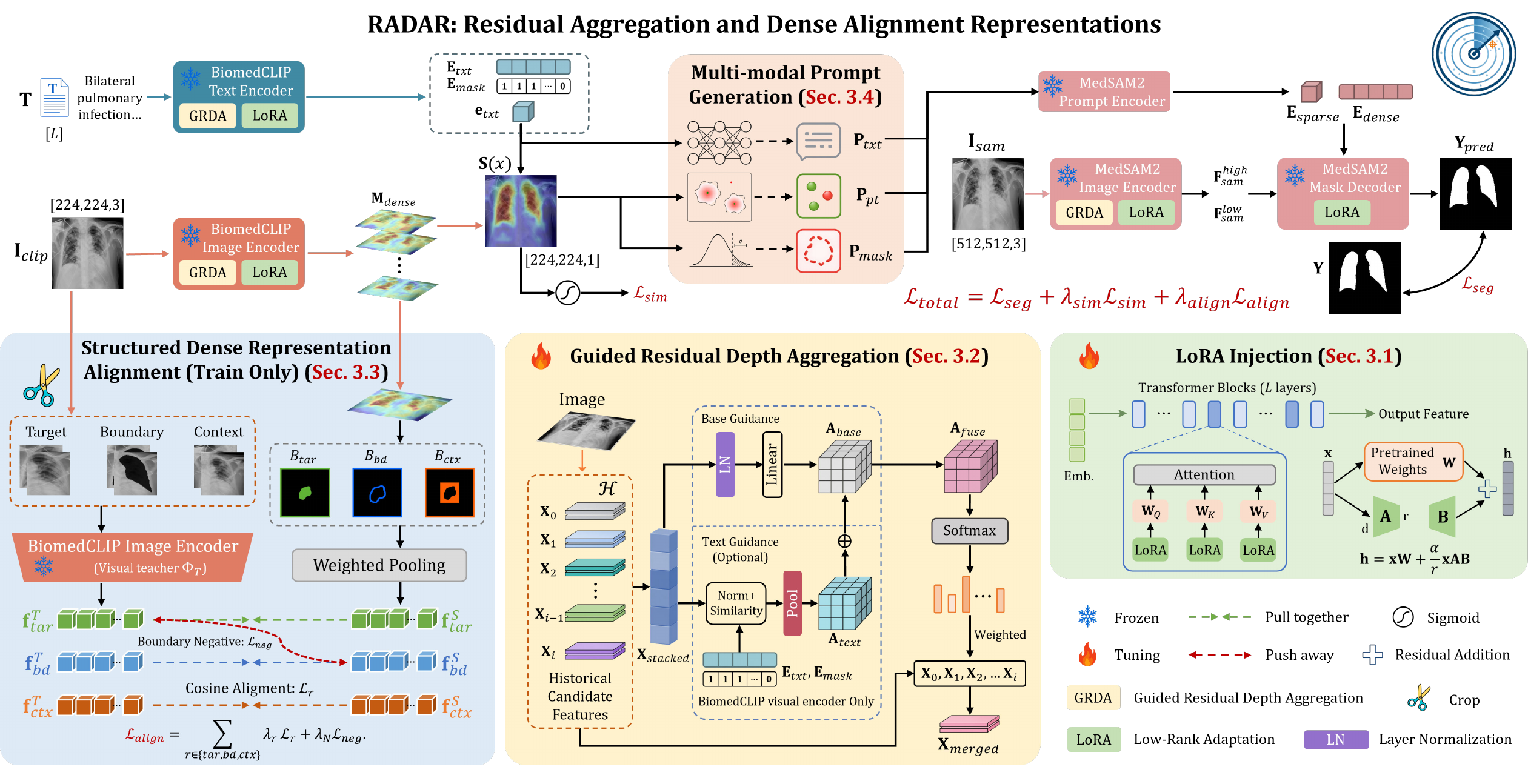}
  \caption{
  Overview of RADAR. GRDA performs text-conditioned cross-layer routing, SDRA provides training-only dense alignment, and MPG converts cross-modal responses into sparse and dense prompts for MedSAM2. The SDRA teacher is removed at inference.
  }
  \label{fig:method}
\end{figure}

\subsection{Overview}

As illustrated in Figure~\ref{fig:method}, given a medical image 
$\mathbf{I} \in \mathbb{R}^{H \times W \times C}$ and a text description 
$\mathbf{T} \in \mathbb{R}^{L}$, we resize the image into 
$\mathbf{I}_{clip} \in \mathbb{R}^{224 \times 224 \times 3}$ for BiomedCLIP and 
$\mathbf{I}_{sam} \in \mathbb{R}^{512 \times 512 \times 3}$ for MedSAM2. 
The BiomedCLIP text branch encodes $\mathbf{T}$ into token-level features 
$\mathbf{E}_{txt} \in \mathbb{R}^{L \times 768}$, a text mask 
$\mathbf{E}_{mask} \in \{0, 1\}^L$ , and a global text feature 
$\mathbf{e}_{txt} \in \mathbb{R}^{512}$. 
Meanwhile, the BiomedCLIP visual branch extracts dense visual features 
$\mathbf{M}_{dense} \in \mathbb{R}^{768 \times 14 \times 14}$ from $\mathbf{I}_{clip}$, 
and the MedSAM2 image encoder extracts multi-scale visual features 
$\mathbf{F}_{sam}^{low}$ and $\mathbf{F}_{sam}^{high}$ from $\mathbf{I}_{sam}$. Based on $\mathbf{e}_{txt}$, we generate a text prompt $\mathbf{P}_{txt}$. 
Point and mask prompts, denoted as $\mathbf{P}_{pt}$ and $\mathbf{P}_{mask}$, are automatically mined from the similarity map between $\mathbf{M}_{dense}$ and $\mathbf{e}_{txt}$. 
The resulting composite prompt set 
$\{\mathbf{P}_{txt}, \mathbf{P}_{pt}, \mathbf{P}_{mask}\}$ is fed into the MedSAM2 prompt encoder to produce sparse embeddings $\mathbf{E}_{sparse}$ and dense embeddings $\mathbf{E}_{dense}$. 
Finally, the MedSAM2 decoder combines these prompt embeddings with high- and low-resolution image features to generate the final segmentation mask $\mathbf{Y}_{pred}$.

\paragraph{LoRA injection.}

To preserve the generalization ability of the pre-trained backbones while enabling parameter-efficient adaptation, we freeze the pretrained backbone weights of BiomedCLIP and MedSAM2 and insert Low-Rank Adaptation (LoRA) modules into selected Transformer layers. 
For a frozen weight matrix $\mathbf{W} \in \mathbb{R}^{d_{in} \times d_{out}}$, the forward computation is reformulated as
\begin{equation}
    \mathbf{h}
    =
    \mathbf{x}\mathbf{W}
    +
    \frac{\alpha}{r}\mathbf{x}\mathbf{A}\mathbf{B},
    \label{eq:lora}
\end{equation}

where $\mathbf{A} \in \mathbb{R}^{d_{in} \times r}$ and 
$\mathbf{B} \in \mathbb{R}^{r \times d_{out}}$ are trainable low-rank matrices, 
$r \ll \min(d_{in}, d_{out})$ is the rank, and $\alpha$ is the scaling factor. 
In practice, LoRA is selectively injected into different Transformer layer ranges to control the adaptation capacity and parameter budget.

\subsection{Guided Residual Depth Aggregation}

To enable text-conditioned selection of historical visual evidence and alleviate feature dilution during deep propagation, we propose Guided Residual Depth Aggregation (GRDA).
For the current layer feature $\mathbf{X}_i$, we collect its historical features as 
$\mathcal{H}_i=\{\mathbf{X}_j\}_{j\leq i}$ and stack them along the depth dimension:
$\mathbf{X}_{stacked}\in\mathbb{R}^{K\times B\times N\times C}$, where $K$ is the number of candidate layers. 
Given an aggregation score $\mathbf{A}\in\mathbb{R}^{K\times B\times N}$, the normalized depth-wise weights and merged feature are computed as
\begin{equation}
    \boldsymbol{\alpha}
    =
    \operatorname{Softmax}
    \left(
    \frac{\mathbf{A}}{\tau},
    \operatorname{dim}=0
    \right),
    \quad
    \mathbf{X}_{merged}
=
\sum_{j\leq i}
\boldsymbol{\alpha}_{j}
\odot
\mathbf{X}_{j}
    \label{eq:grda_merge}
\end{equation}

where $\tau$ is a temperature parameter. The aggregated feature is injected through a learnable residual scale:
\begin{equation}
    \mathbf{X}_{out}
    =
    \mathbf{X}_{i}
    +
    \gamma
    \left(
    \mathbf{X}_{merged}
    -
    \mathbf{X}_{i}
    \right).
    \label{eq:grda_output}
\end{equation}

We initialize $\gamma$ to $0$, so that GRDA starts as an identity mapping and preserves the pretrained representations at the beginning of training.

\paragraph{Base aggregation score.}

For the BiomedCLIP text encoder and the MedSAM2 image encoder, GRDA computes feature-intrinsic aggregation scores without cross-modal intervention:
\begin{equation}
    \mathbf{A}_{base}
    =
    \operatorname{Squeeze}\left(\operatorname{Linear}_{C \to 1}
    \left(
    \operatorname{LayerNorm}
    \left(
    \mathbf{X}_{stacked}
    \right)
    \right) \right)
    \in
    \mathbb{R}^{K\times B\times N}.
    \label{eq:grda_base}
\end{equation}

\paragraph{Text-conditioned routing score.}

For the BiomedCLIP visual encoder, we further introduce token-level textual guidance. Specifically, we compute the cosine similarity between the stacked visual features and the text token features $\mathbf{E}_{txt}$, and perform mask-weighted pooling along the text length dimension using the text mask $\mathbf{E}_{mask}$ to exclude padding tokens:
\begin{equation}
    \mathbf{S}_{text}^{(l)} = \cos(\mathbf{X}_{stacked}, \mathbf{E}_{txt}^{(l)}), \quad
    \mathbf{A}_{text} = \frac{\sum_{l=1}^L \mathbf{S}_{text}^{(l)} \cdot \mathbf{E}_{mask}^{(l)}}{\sum_{l=1}^L \mathbf{E}_{mask}^{(l)}}.
    \label{eq:grda_text}
\end{equation}

Finally, the fused aggregation score is obtained by additive fusion: $\mathbf{A}_{fuse}=\mathbf{A}_{base}+\mathbf{A}_{text}$.

\subsection{Structured Dense Representation Alignment}

The BiomedCLIP visual encoder is originally optimized for global image-text retrieval, where deep features may lose local boundaries and spatial details required for medical segmentation. Moreover, its dense grid features lack sufficient local discriminability to separate target regions, blurred boundaries, and similar backgrounds. To address this, we modify the last ViT block and introduce Structured Dense Representation Alignment (SDRA) to enhance the local discriminability of dense representations.

\paragraph{Dense feature extraction modification.}

Following~\citet{wu2024clipself} and~\citet{zhou2022extract}, we modify the last ViT block of the BiomedCLIP visual encoder by replacing the original attention output with the projected Value-branch representation:
\begin{equation}
    \mathbf{y}
    =
    \mathbf{x}
    +
    \operatorname{Proj}
    \left(
    \operatorname{Emb}_v(\mathbf{x})
    \right),
    \quad
    \mathbf{z}
    =
    \mathbf{y}
    +
    \operatorname{FFN}(\mathbf{y}),
    \label{eq:dense_feature_modification}
\end{equation}

where $\mathbf{x}$ denotes the input visual tokens, $\operatorname{Emb}_v(\cdot)$ denotes the Value-branch mapping, and $\operatorname{Proj}(\cdot)$ is the output projection. The output $\mathbf{z}$ is reshaped into a spatial grid to obtain the dense visual feature map $\mathbf{M}_{dense}$.

\paragraph{Structured triplet construction.}

Given the ground-truth mask $\mathbf{Y}$ during training, we first compute its minimum bounding box $B_{orig}$. Then, three structured regions are constructed by box expansion: $B_{tar}=\operatorname{Expand}(B_{orig},v_{tar})$, $B_{bd}=\operatorname{Expand}(B_{tar},v_{bd})$, and $B_{ctx}=\operatorname{Expand}(B_{tar},v_{ctx})$. Here, $B_{tar}$ denotes the target-centered region, $B_{bd}$ covers blurred boundaries and transition regions, and $B_{ctx}$ captures surrounding context or similar background structures. These nested structured regions provide region-level supervision that better matches common segmentation errors in medical images.

\paragraph{Crop-level teacher and student representations.}

For each region $r\in\{tar,bd,ctx\}$, a frozen BiomedCLIP image encoder $\Phi_T$ extracts the crop-level teacher anchor:
$\mathbf{f}_r^T=\Phi_T(\operatorname{Crop}(\mathbf{I}_{clip},B_r))$.
Meanwhile, the student representation is obtained by mask-weighted pooling over $\mathbf{M}_{dense}$ using the corresponding spatial mask $\mathbf{M}_r$ associated with bounding box $B_r$:
$\mathbf{f}_r^S = \operatorname{Pool}_{mask}(\mathbf{M}_{dense}, \mathbf{M}_r)$.
The regional alignment loss is defined as
\begin{equation}
    \mathcal{L}_r
    =
    1
    -
    \left\langle
    \frac{\mathbf{f}_r^S}{\|\mathbf{f}_r^S\|_2+\epsilon},
    \frac{\mathbf{f}_r^T}{\|\mathbf{f}_r^T\|_2+\epsilon}
    \right\rangle,
    \quad
    r\in\{tar,bd,ctx\}.
    \label{eq:regional_alignment_loss}
\end{equation}

\paragraph{Boundary negative rejection.}
Since boundary regions may contain both foreground and background information, they can be incorrectly absorbed into the target representation. We therefore introduce a boundary negative rejection term:
\begin{equation}
    \mathcal{L}_{neg}
    =
    \operatorname{ReLU}
    \left(
    \cos(\mathbf{f}_{bd}^S,\mathbf{f}_{tar}^T)
    -
    \cos(\mathbf{f}_{tar}^S,\mathbf{f}_{tar}^T)
    \right).
    \label{eq:boundary_negative_rejection}
\end{equation}

This term encourages the boundary representation to be less similar to the target teacher anchor than the target student representation.

The final structured alignment loss is
\begin{equation}
    \mathcal{L}_{align}
    =
    \sum_{r\in\{tar,bd,ctx\}}
    \lambda_r\mathcal{L}_r
    +
    \lambda_N\mathcal{L}_{neg},
    \label{eq:structured_alignment_loss}
\end{equation}

where $\lambda_r$ and $\lambda_N$ denote the weights of the regional alignment and boundary negative rejection terms, respectively.

\subsection{Multi-modal Prompt Generation}

To convert BiomedCLIP cross-modal semantics into spatial control signals for MedSAM2, we design Multi-modal Prompt Generation (MPG), which produces a composite prompt set 
$\mathcal{P}=\{\mathbf{P}_{txt},\mathbf{P}_{pt},\mathbf{P}_{mask}\}$.

\paragraph{Text prompt generation.}

The global text embedding $\mathbf{e}_{txt}$ from the BiomedCLIP text branch encodes the target category and medical semantics. We project it into the MedSAM2 prompt space with a two-layer MLP:
$\mathbf{P}_{txt}=\mathbf{W}_2\,\operatorname{GELU}(\mathbf{W}_1\mathbf{e}_{txt})$, where $\mathbf{W}_1$ and $\mathbf{W}_2$ are learnable projection matrices.

\paragraph{Mask and point prompt generation.}

To align the dimensionalities, the dense visual features $\mathbf{M}_{dense}$ are first mapped to the shared semantic space using the frozen BiomedCLIP visual projection layer. We then compute the cross-modal cosine similarity map $\mathbf{S}(\mathbf{x})$ between the projected dense features and the global text feature $\mathbf{e}_{txt}$. For mask prompting, the similarity response is normalized and resized to the mask-prompt resolution required by MedSAM2, yielding $\mathbf{P}_{mask}$. For point prompting, $\mathbf{S}(\mathbf{x})$ is resized to the MedSAM2 input resolution and processed by adaptive quantile thresholding to obtain high-confidence foreground regions. Connected components are then extracted from these regions. Positive points are sampled from eroded high-confidence foreground cores, while negative points are sampled from surrounding hard-background regions. The resulting point prompt set is
$\mathbf{P}_{pt}=\{(\mathbf{p}_i^+,1)\}\cup\{(\mathbf{p}_j^-,0)\}$, where labels $1$ and $0$ denote foreground and background, respectively. Detailed point prompt generation settings are provided in Appendix~\ref{app:mpg_details}.

\paragraph{Training objective.}

The final MedSAM2 prediction $\mathbf{Y}_{pred}$ is supervised by a BCE-Dice segmentation loss:
$\mathcal{L}_{seg}=0.5\mathcal{L}_{BCE}(\mathbf{Y}_{pred},\mathbf{Y})+0.5\mathcal{L}_{Dice}(\mathbf{Y}_{pred},\mathbf{Y})$. We convert the similarity map $\mathbf{S}(\mathbf{x})$ into a probability map using a sigmoid function and supervise it with the same BCE-Dice formulation as an auxiliary loss $\mathcal{L}_{sim}$. Detailed loss definitions are provided in Appendix~\ref{app:loss_details}.  Together with the structured dense representation alignment loss $\mathcal{L}_{align}$, the total objective is
\begin{equation}
    \mathcal{L}_{total}
    =
    \mathcal{L}_{seg}
    +
    \lambda_{sim}\mathcal{L}_{sim}
    +
    \lambda_{align}\mathcal{L}_{align},
    \label{eq:total_loss}
\end{equation}

where $\lambda_{sim}$ and $\lambda_{align}$ control the weights of similarity-map supervision and structured alignment, respectively.

\section{Experiments}

\subsection{Experimental settings}

\paragraph{Datasets and metrics.}

We evaluate RADAR on three public medical image segmentation datasets, including two chest infection segmentation datasets, QaTa-COV19~\citep{Degerli2022Osegnet} and MosMedData+~\citep{morozov2020mosmeddata,hofmanninger2020automatic}, and one colonoscopy polyp segmentation dataset, Kvasir~\citep{Jha2020Kvasir}. 
For QaTa-COV19 and MosMedData+, we follow the data splits and text annotations provided by LViT~\citep{li2024LViT}. 
For Kvasir, we use the text annotations and experimental setting from MAdapter~\citep{Zha_MAdapter_MICCAI2024}, and split the original training set into training and validation subsets with a 9:1 ratio, while keeping the standard test set fixed. 
All compared methods are retrained and evaluated under the same data splits. 
We report Dice, mean Intersection over Union (mIoU), 95th percentile Hausdorff Distance (HD95), and Average Symmetric Surface Distance (ASSD). 
Detailed dataset statistics and metric definitions are provided in Appendix~\ref{app:datasets_metrics}.

\paragraph{Implementation details.}

All experiments are conducted on Ubuntu 25.04 with an Intel Core i9-14900KF CPU and a single NVIDIA GeForce RTX 5090 D GPU using PyTorch 2.9.0 and CUDA 12.8. 
The BiomedCLIP and MedSAM2 backbones are frozen during training. 
For LoRA, we set the rank to $r=8$ and the scaling factor to $\alpha=1.0$. 
The numerical stability term is set to $\epsilon=10^{-6}$. 
For structured region alignment, we set $\lambda_{tar}=0.5$ and $\lambda_{bd}=\lambda_{ctx}=\lambda_N=0.2$. 
For the total objective, we set $\lambda_{sim}=0.2$ and $\lambda_{align}=0.2$. 
We train the model with AdamW using an initial learning rate of $3\times10^{-4}$, a batch size of $8$, CosineAnnealingWarmRestarts scheduling, and early stopping.

\subsection{Comparison with existing methods}

\paragraph{Quantitative comparison.}

As shown in Table~\ref{tab:quantitative_comparison}, RADAR achieves the best Dice and mIoU on QaTa-COV19, MosMedData+, and Kvasir. Compared with the strongest pure visual baselines, it improves Dice by 6.65, 10.12, and 5.96 percentage points, respectively. Against the best V-L methods, RADAR further improves Dice/mIoU by 0.84/1.01 and 1.53/2.04 percentage points on MosMedData+ and Kvasir. Compared with the best external V-L+S baselines, it improves Dice by 2.65, 3.01, and 2.35 percentage points while maintaining more stable boundary quality. RADAR requires only 0.74M trainable parameters and 46.66G FLOPs, demonstrating strong task-adaptation efficiency with moderate forward computation. Cross-backbone results further support the synergy between medical-domain semantic pretraining and a medical promptable decoder. Robustness to incomplete, noisy, synonymous, and incorrect text is further evaluated in Appendix~\ref{app:text_robustness}.
 
\begin{table}[htbp]
\caption{Quantitative comparison across three datasets. Dice and mIoU are reported in percentages (\%), while HD95/ASSD are reported in pixels. Params denotes the number of trainable parameters. ``Visual'', ``V-L'', and ``V-L+S'' denote pure vision models, vision-language models, and VLM-SAM cascades, respectively. BioCLIP denotes BiomedCLIP. CLIP-SAM2, BioCLIP-SAM2, and CLIP-MedSAM2 correspond to RADAR backbone variants equipped with the full GRDA+SDRA configuration. Best and second-best results are \textbf{bolded} and \underline{underlined}, respectively.}
\label{tab:quantitative_comparison}
\centering
\scriptsize
\setlength{\tabcolsep}{1.5pt}
\renewcommand{\arraystretch}{1.0}
\resizebox{\linewidth}{!}{
\begin{tabular}{llcc|cccc|cccc|cccc}
\toprule
\multirow{2}{*}{Method} & \multirow{2}{*}{Type}
& \multirow{2}{*}{\makecell{Trainable\\Params $\downarrow$}}
& \multirow{2}{*}{FLOPs $\downarrow$}
& \multicolumn{4}{c|}{QaTa-COV19}
& \multicolumn{4}{c|}{MosMedData+}
& \multicolumn{4}{c}{Kvasir} \\
\cmidrule(lr){5-8} \cmidrule(lr){9-12} \cmidrule(lr){13-16}
& & &
& Dice $\uparrow$ & mIoU $\uparrow$ & HD95 $\downarrow$ & ASSD $\downarrow$
& Dice $\uparrow$ & mIoU $\uparrow$ & HD95 $\downarrow$ & ASSD $\downarrow$
& Dice $\uparrow$ & mIoU $\uparrow$ & HD95 $\downarrow$ & ASSD $\downarrow$ \\
\midrule

U-Net~\citep{Ronneberger2015UNet} & Visual & 31.04M & 41.91G
& 76.95 & 66.97 & 30.75 & 8.68
& 40.57 & 30.41 & 42.55 & 14.81
& 54.82 & 41.38 & 54.33 & 20.06 \\

UNet++~\citep{Zhou2018UNet++} & Visual & 9.16M & 26.72G
& 76.60 & 66.29 & 35.00 & 9.54
& 65.35 & 50.78 & 23.28 & 5.81
& 61.74 & 49.13 & 55.25 & 19.20 \\

Swin-UNet~\citep{Cao2023SwinUnet} & Visual & 27.17M & \textbf{5.92G}
& 77.47 & 68.05 & 26.28 & 7.55
& 63.98 & 50.61 & 22.45 & 5.89
& 83.90 & 75.98 & 19.51 & 6.37 \\

TransUNet~\citep{Chen2024TransUNet} & Visual & 105.28M & 49.36G
& 77.38 & 67.27 & 30.21 & 8.01
& 63.52 & 49.26 & 23.83 & 5.81
& 86.19 & 78.50 & 23.28 & 6.64 \\

\midrule

LGMS~\citep{Zhong2023Ariadne} & V-L & 43.95M & 11.18G
& 83.33 & 74.33 & 15.78 & 4.20
& 71.83 & 58.22 & \underline{13.74} & \underline{3.29}
& \underline{90.62} & 84.45 & 16.74 & 4.55 \\

LViT~\citep{li2024LViT} & V-L & 39.93M & 27.08G
& 83.21 & 74.43 & 17.43 & 4.73
& 73.97 & 60.89 & 18.55 & 4.40
& 88.14 & 81.07 & 21.56 & 5.59 \\

MMI-UNet~\citep{Bui2024Visual} & V-L & 56.18M & \underline{11.04G}
& 83.86 & 75.20 & 16.10 & 4.35
& 70.80 & 57.23 & 14.89 & 4.14
& 90.42 & 84.61 & \textbf{14.16} & \textbf{3.82} \\

LAVT~\citep{yang2025LAVT} & V-L & 189.73M & 49.69G
& 83.81 & \underline{75.27} & \textbf{14.07} & \textbf{3.88}
& 67.00 & 53.28 & 16.81 & 4.24
& 89.54 & 83.46 & \underline{14.68} & 4.66 \\

RecLMIS~\citep{huang2025cross} & V-L & 69.44M & 26.69G
& \underline{83.87} & 75.16 & 15.36 & 4.19
& \underline{74.63} & \underline{61.42} & 18.88 & 4.31
& 84.12 & 75.78 & 23.76 & 6.42 \\

MTGT~\citep{ZHAO2026MTGT} & V-L & 56.73M & 128.56G
& 77.22 & 67.71 & 27.24 & 8.12
& 68.01 & 54.22 & 16.78 & 4.09
& 71.76 & 62.40 & 36.98 & 11.86 \\

HiMix~\citep{Hwang_2026_WACV} & V-L & 44.73M & 11.27G
& 82.56 & 73.25 & 18.89 & 5.04
& 72.57 & 58.96 & 14.64 & 3.91
& 89.94 & 84.09 & 17.85 & 4.60 \\

\midrule

DIT~\citep{huang2024deep} & V-L+S & 103.38M & 92.82G
& 79.99 & 70.40 & 28.79 & 7.81
& 72.22 & 59.00 & 15.19 & 3.58
& 89.80 & 83.97 & 16.70 & 5.01 \\

LGA~\citep{Hu_LGA_MICCAI2024} & V-L+S & 97.91M & 381.18G
& 81.47 & 72.22 & 21.90 & 6.27
& 72.46 & 58.92 & 20.09 & 4.76
& 86.37 & 78.60 & 23.15 & 6.77 \\

\midrule

CLIP-SAM2 & V-L+S & \underline{1.13M} & 252.66G
& 81.83 & 72.33 & 16.97 & 5.22
& 74.58 & 61.28 & 14.71 & 3.74
& 88.25 & 82.26 & 16.75 & 4.91 \\

BioCLIP-SAM2 & V-L+S & \textbf{0.74M} & 254.55G
& 81.62 & 71.94 & 18.81 & 5.47
& 73.15 & 59.35 & 16.58 & 4.48
& 90.02 & 84.60 & 17.82 & 4.95 \\

CLIP-MedSAM2 & V-L+S & \underline{1.13M} & 93.85G
& 83.35 & 74.36 & 15.97 & 4.77
& 73.90 & 60.43 & 14.83 & 3.83
& 90.27 & \underline{84.80} & 16.37 & 4.51 \\

RADAR (Ours) & V-L+S & \textbf{0.74M} & 46.66G
& \textbf{84.12} & \textbf{75.32} & \underline{14.27} & \underline{4.09}
& \textbf{75.47} & \textbf{62.43} & \textbf{13.16} & \textbf{3.26}
& \textbf{92.15} & \textbf{86.65} & 14.79 & \underline{3.91} \\

\bottomrule
\end{tabular}
}
\end{table}

\paragraph{Qualitative analysis.}

\begin{figure}[htbp]
  \centering
  \includegraphics[width=0.95\linewidth]{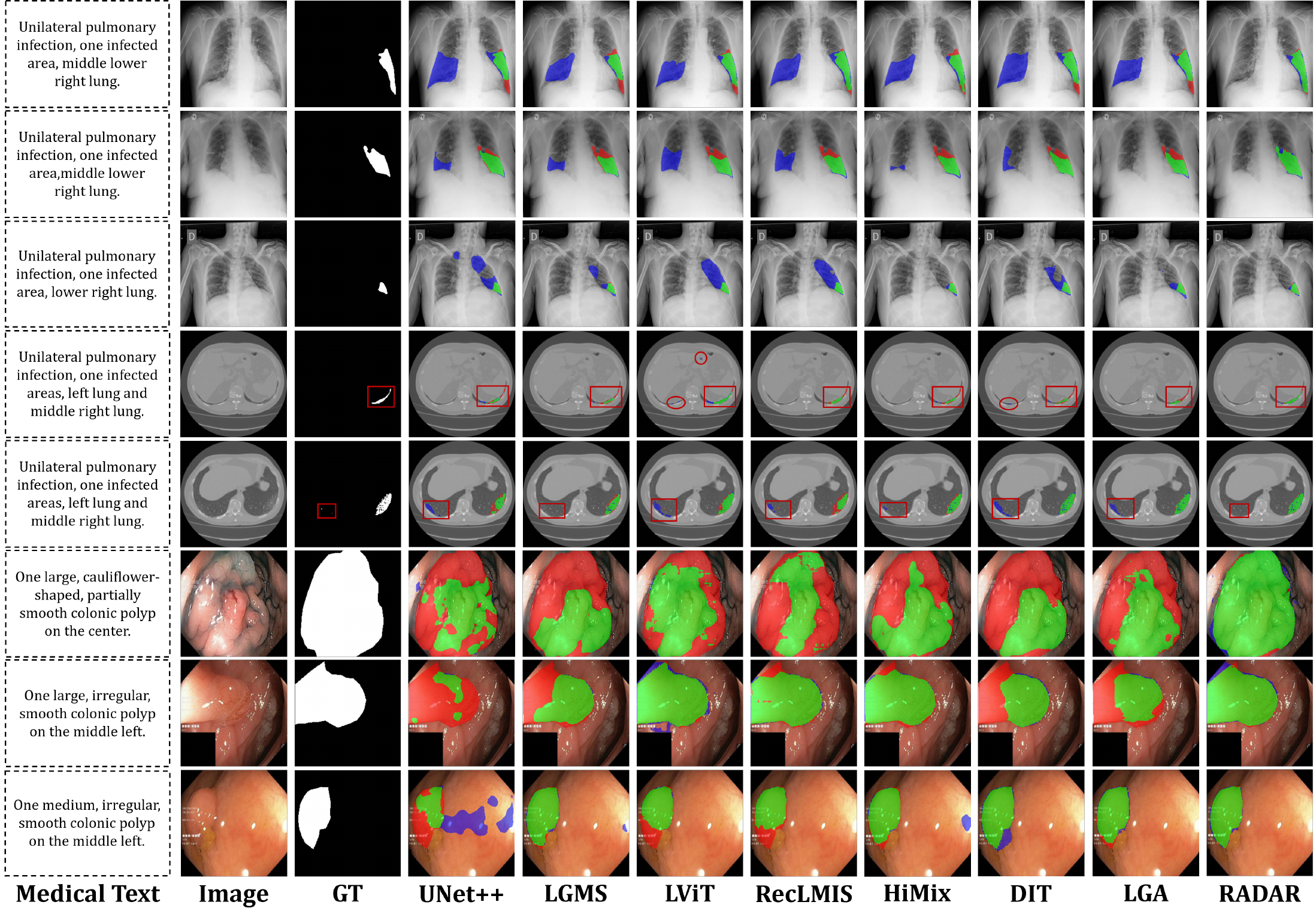}
  \caption{
  Qualitative comparison on representative QaTa-COV19, MosMedData+, and Kvasir samples covering low-contrast lesions, small targets, and ambiguous boundaries. Green, red, and blue indicate true positive, false negative, and false positive pixels, respectively. RADAR shows improved target localization, background suppression, and boundary recovery across these challenging cases.
  }
  \label{fig:qualitative_comparison}
\end{figure}

The qualitative results further support these observations. As shown in Figure~\ref{fig:qualitative_comparison}, pure visual methods generally suffer from unstable localization and boundary deviations, making it difficult to distinguish low-contrast lesions from visually similar anatomical structures, as illustrated in the fourth column. The small-target CT examples in the fourth and fifth rows further show that several methods are prone to missed detections, localization shifts, or background false activations when segmenting tiny lesions. Although vision-language methods introduce textual semantics, they fail to convert such semantics into explicit and reliable spatial prompts, leading to incorrect activations in non-target lung regions, as shown in the first and second rows. Existing V-L+S cascaded methods improve contour quality with SAM-style decoders, but remain highly dependent on the prompts generated by the front-end VLM. Once noisy similarity responses are produced, the errors are propagated to the downstream mask decoding process, as observed in the ninth and tenth columns. In contrast, RADAR exhibits more stable semantic grounding and boundary recovery across different modalities and target scales, validating its advantages in small-target perception, background suppression, and boundary consistency.

\subsection{Ablation studies}

\paragraph{Effect of main component.}

Table~\ref{tab:main_component_ablation} validates the effectiveness of the main components in RADAR. Starting from the MedSAM2-only baseline, progressively adding text, point, and mask prompts consistently improves performance, showing that MPG effectively converts textual semantics into spatial constraints. Adding GRDA on top of Full MPG further raises Dice to 72.78\% and 91.35\% on MosMedData+ and Kvasir, respectively, indicating the benefit of text-conditioned cross-layer routing. Removing SDRA, Dense Mod., or GRDA consistently degrades performance, confirming their complementary roles in local discrimination, spatial detail preservation, and semantic propagation. The full RADAR achieves the best results on both datasets. Additional comparisons with direct and learnable prompt-generation alternatives are provided in Appendix~\ref{app:learnable_mpg}, with further ablations in Appendix~\ref{app:additional_ablation}.

\begin{table}[htbp]
\caption{Main component ablation study on MosMedData+ and Kvasir. Dense Mod. denotes the modification of the last BiomedCLIP ViT block for dense feature extraction.}
\label{tab:main_component_ablation}
\centering
\scriptsize
\setlength{\tabcolsep}{2.5pt}
\renewcommand{\arraystretch}{1.0}
\resizebox{\linewidth}{!}{
\begin{tabular}{lcccccc|cc|cc}
\toprule
\multirow{2}{*}{Variant}
& \multicolumn{3}{c}{MPG}
& \multirow{2}{*}{GRDA}
& \multirow{2}{*}{Dense Mod.}
& \multirow{2}{*}{SDRA}
& \multicolumn{2}{c|}{MosMedData+}
& \multicolumn{2}{c}{Kvasir} \\
\cmidrule(lr){2-4} \cmidrule(lr){8-9} \cmidrule(lr){10-11}
& Text & Point & Mask
& & &
& Dice (\%) $\uparrow$ & mIoU (\%) $\uparrow$
& Dice (\%) $\uparrow$ & mIoU (\%) $\uparrow$ \\
\midrule

MedSAM2-only
& -- & -- & -- & -- & -- & --
& 69.19 & 55.33 & 88.11 & 81.07 \\

+ Text
& $\checkmark$ & -- & -- & -- & -- & --
& 69.86 & 55.68 & 88.73 & 82.17 \\

+ Text + Point
& $\checkmark$ & $\checkmark$ & -- & -- & -- & --
& 71.14 & 57.47 & 89.93 & 83.62 \\

Full MPG
& $\checkmark$ & $\checkmark$ & $\checkmark$ & -- & -- & --
& 71.47 & 57.67 & 90.36 & 84.29 \\

Full MPG + GRDA
& $\checkmark$ & $\checkmark$ & $\checkmark$ & $\checkmark$ & -- & --
& 72.78 & 59.40 & 91.35 & 85.61 \\

\midrule

w/o SDRA
& $\checkmark$ & $\checkmark$ & $\checkmark$ & $\checkmark$ & $\checkmark$ & --
& 73.71 & 60.22 & 91.63 & \underline{86.25} \\

w/o Dense Mod.
& $\checkmark$ & $\checkmark$ & $\checkmark$ & $\checkmark$ & -- & $\checkmark$
& \underline{75.29} & \underline{62.21} & \underline{91.77} & 86.24 \\

w/o GRDA
& $\checkmark$ & $\checkmark$ & $\checkmark$ & -- & $\checkmark$ & $\checkmark$
& 74.63 & 61.30 & 90.84 & 85.28 \\

RADAR (Ours)
& $\checkmark$ & $\checkmark$ & $\checkmark$ & $\checkmark$ & $\checkmark$ & $\checkmark$
& \textbf{75.47} & \textbf{62.43} & \textbf{92.15} & \textbf{86.65} \\

\bottomrule
\end{tabular}
}
\end{table}

\paragraph{Effect of GRDA and SDRA.}

Table~\ref{tab:grda_sdra_ablation} validates the transferability and complementarity of GRDA and SDRA across different backbone combinations. On CLIP-SAM2, BioCLIP-SAM2, and CLIP-MedSAM2, introducing either GRDA or SDRA consistently improves performance over the baseline without both modules, indicating that semantic-guided cross-layer aggregation and structured dense alignment are not tied to a specific backbone. Furthermore, jointly incorporating GRDA and SDRA achieves the best Dice and mIoU across all backbone combinations, demonstrating that the two modules complement each other in stabilizing semantic propagation and enhancing local dense discriminability, thereby improving medical semantic grounding.

\begin{table}[htbp]
\caption{Cross-backbone ablation study of GRDA and SDRA on MosMedData+ and Kvasir. BioCLIP denotes BiomedCLIP.}
\label{tab:grda_sdra_ablation}
\centering
\scriptsize
\setlength{\tabcolsep}{2.5pt}
\renewcommand{\arraystretch}{1.0}
\resizebox{\linewidth}{!}{
\begin{tabular}{llcc|cccc|cccc}
\toprule
\multirow{2}{*}{Backbone} & \multirow{2}{*}{Variant}
& \multirow{2}{*}{GRDA} & \multirow{2}{*}{SDRA}
& \multicolumn{4}{c|}{MosMedData+}
& \multicolumn{4}{c}{Kvasir} \\
\cmidrule(lr){5-8} \cmidrule(lr){9-12}
& & &
& Dice $\uparrow$ & mIoU $\uparrow$ & HD95 $\downarrow$ & ASSD $\downarrow$
& Dice $\uparrow$ & mIoU $\uparrow$ & HD95 $\downarrow$ & ASSD $\downarrow$ \\
\midrule

\multirow{4}{*}{CLIP-SAM2}
& w/o GRDA \& SDRA & $\times$ & $\times$
& 71.86 & 57.90 & 19.06 & 5.06
& 86.11 & 79.08 & 21.73 & 6.25 \\
& GRDA only & $\checkmark$ & $\times$
& \underline{73.84} & \underline{60.35} & \underline{17.41} & \underline{4.36}
& 87.57 & 81.03 & \underline{19.62} & 5.58 \\
& SDRA only & $\times$ & $\checkmark$
& 73.01 & 59.33 & 17.91 & 4.72
& \underline{87.79} & \underline{81.21} & 19.63 & \underline{5.47} \\
& GRDA + SDRA & $\checkmark$ & $\checkmark$
& \textbf{74.58} & \textbf{61.28} & \textbf{14.71} & \textbf{3.74}
& \textbf{88.25} & \textbf{82.26} & \textbf{16.75} & \textbf{4.91} \\
\midrule

\multirow{4}{*}{BioCLIP-SAM2}
& w/o GRDA \& SDRA & $\times$ & $\times$
& 71.70 & 57.85 & 18.71 & 5.08
& 88.91 & 82.56 & 21.46 & 6.01 \\
& GRDA only & $\checkmark$ & $\times$
& 72.10 & 58.25 & \underline{17.04} & \underline{4.68}
& \underline{89.56} & \underline{83.69} & 19.48 & 5.46 \\
& SDRA only & $\times$ & $\checkmark$
& \underline{72.47} & \underline{58.91} & 17.28 & 4.93
& 89.45 & 83.60 & \underline{19.07} & \underline{5.31} \\
& GRDA + SDRA & $\checkmark$ & $\checkmark$
& \textbf{73.15} & \textbf{59.35} & \textbf{16.58} & \textbf{4.48}
& \textbf{90.02} & \textbf{84.60} & \textbf{17.82} & \textbf{4.95} \\
\midrule

\multirow{4}{*}{CLIP-MedSAM2}
& w/o GRDA \& SDRA & $\times$ & $\times$
& 72.56 & 58.99 & 16.03 & 4.42
& 88.82 & 82.15 & 21.13 & 6.02 \\
& GRDA only & $\checkmark$ & $\times$
& \underline{73.31} & 59.75 & 16.95 & 4.23
& \underline{89.84} & \underline{83.95} & \textbf{15.35} & \textbf{4.41} \\
& SDRA only & $\times$ & $\checkmark$
& 73.23 & \underline{59.80} & \underline{15.79} & \underline{4.05}
& 89.75 & 83.58 & 19.83 & 5.47 \\
& GRDA + SDRA & $\checkmark$ & $\checkmark$
& \textbf{73.90} & \textbf{60.43} & \textbf{14.83} & \textbf{3.83}
& \textbf{90.27} & \textbf{84.80} & \underline{16.37} & \underline{4.51} \\

\bottomrule
\end{tabular}
}
\end{table}

\paragraph{Ablation study on GRDA variants.}

Table~\ref{tab:grda_ablation} examines two internal design choices of GRDA: aggregation score design and branch placement. For score design, the Base and Text scores are individually effective, while their additive fusion achieves the best Dice/mIoU on both datasets. Text-only yields slightly better boundary metrics on MosMedData+, indicating complementary structural and semantic cues and making Base+Text the most balanced choice overall. For branch placement, restricting GRDA to S-I or removing either B-V or B-T degrades performance, whereas jointly applying it to S-I, B-V, and B-T performs best overall. These results support feature-intrinsic cross-layer aggregation across branches together with text-conditioned routing in the BiomedCLIP visual branch.

\begin{table}[htbp]
\caption{
Ablation of GRDA score design and branch placement on MosMedData+ and Kvasir. S-I, B-V, and B-T denote the MedSAM2 image, BiomedCLIP visual, and text encoders, respectively. Base and Text denote feature-intrinsic and text-conditioned scores in B-V.
}
\label{tab:grda_ablation}
\centering
\scriptsize
\setlength{\tabcolsep}{2.0pt}
\renewcommand{\arraystretch}{1.0}
\resizebox{\linewidth}{!}{
\begin{tabular}{lll|cccc|cccc}
\toprule
\multirow{2}{*}{Variant}
& \multirow{2}{*}{Score}
& \multirow{2}{*}{GRDA branches}
& \multicolumn{4}{c|}{MosMedData+}
& \multicolumn{4}{c}{Kvasir} \\
\cmidrule(lr){4-7} \cmidrule(lr){8-11}
& &
& Dice $\uparrow$ & mIoU $\uparrow$ & HD95 $\downarrow$ & ASSD $\downarrow$
& Dice $\uparrow$ & mIoU $\uparrow$ & HD95 $\downarrow$ & ASSD $\downarrow$ \\
\midrule

\multicolumn{11}{l}{\textit{Aggregation score design}} \\

Base-only
& Base
& S-I, B-V, B-T
& 75.13 & 62.01 & 13.37 & 3.28
& 91.75 & 86.15 & 17.60 & 4.53 \\

Multiplicative
& Base $\times$ Text
& S-I, B-V, B-T
& 75.35 & \underline{62.32} & 13.22 & 3.37
& 91.90 & 86.21 & 16.69 & 4.24 \\

Text-only
& Text
& S-I, B-V, B-T
& 75.22 & 62.11 & \textbf{12.83} & \textbf{3.20}
& \underline{92.00} & \underline{86.44} & \underline{15.48} & \underline{3.93} \\

\midrule

\multicolumn{11}{l}{\textit{GRDA branch placement}} \\

S-I only
& --
& S-I
& 74.82 & 61.72 & 13.27 & 3.41
& 91.00 & 84.98 & 19.49 & 4.72 \\

w/o B-V
& --
& S-I, B-T
& 75.30 & 62.12 & 13.32 & 3.36
& 91.33 & 85.45 & 18.76 & 4.81 \\

w/o B-T
& Base $+$ Text
& S-I, B-V
& \underline{75.38} & 62.28 & 13.92 & 3.40
& 91.84 & 86.14 & 17.23 & 4.37 \\

\midrule

RADAR (Ours)
& Base $+$ Text
& S-I, B-V, B-T
& \textbf{75.47} & \textbf{62.43} & \underline{13.16} & \underline{3.26}
& \textbf{92.15} & \textbf{86.65} & \textbf{14.79} & \textbf{3.91} \\

\bottomrule
\end{tabular}
}
\end{table}

\paragraph{Visualization analysis of SDRA.}

\begin{figure}[htbp]
  \centering
  \includegraphics[width=0.9\linewidth]{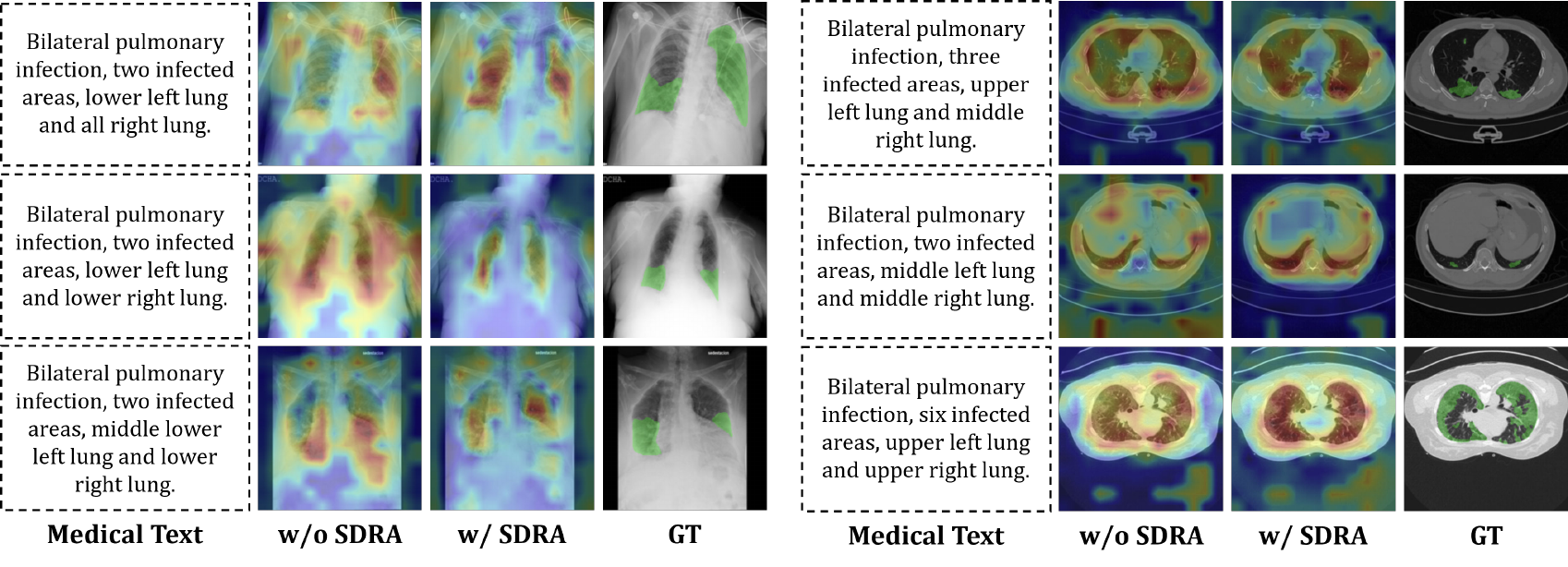}
  \caption{
  Qualitative visualization of SDRA. The columns show the medical text, BiomedCLIP dense features without SDRA, BiomedCLIP dense features with SDRA, and ground-truth masks.
  }
  \label{fig:sdra_visualization}
\end{figure}

Figure~\ref{fig:sdra_visualization} illustrates the effect of SDRA on the dense representations of BiomedCLIP on QaTa-COV19 and MosMedData+. The second and third columns show the last-layer dense outputs of the BiomedCLIP visual encoder without and with SDRA, respectively. Without SDRA, the dense responses are scattered over non-target lung regions, bone structures, and visually similar anatomical backgrounds, indicating that the original BiomedCLIP dense features mainly capture coarse text-related semantics but lack fine-grained local discriminability. After introducing SDRA, the responses become more concentrated around the true lesion regions, with irrelevant background activations clearly suppressed and better spatial consistency with the ground truth. These results demonstrate that SDRA effectively enhances the local discriminability of dense representations, providing a more reliable spatial basis for automatic prompt generation and fine-grained segmentation. More extended visual evidence and discussion of SDRA are provided in Appendix~\ref{app:sdra_visualization}.

\section{Conclusion}

We propose RADAR, a parameter-efficient framework for text-guided medical image segmentation. RADAR decouples the semantic anchoring capability of BiomedCLIP from the mask decoding ability of MedSAM2 and enables low-cost adaptation via LoRA. GRDA preserves target-relevant visual evidence through text-conditioned cross-layer routing, SDRA improves local dense-feature discriminability through structured regional alignment, and MPG converts enhanced cross-modal responses into executable text, point, and mask prompts. Together, these components form a continuous semantic-to-spatial pathway from textual guidance to pixel-level segmentation. Experiments across multiple medical segmentation benchmarks demonstrate that RADAR achieves strong and consistent performance with only 0.74M trainable parameters, highlighting its effectiveness for parameter-efficient collaboration between medical VLMs and promptable segmentation models.

\section{Limitations}

Despite its strong performance, RADAR still has two main limitations. First, it relies on the pretrained priors of BiomedCLIP and MedSAM2, and current validation mainly focuses on 2D tasks without native 3D, multi-organ, or simultaneous multiclass evaluation. Second, MPG still relies on heuristic thresholding, connected-component analysis, and morphological operations, which may be unstable under low contrast, complex targets, or inaccurate and incomplete text. Future work will explore learnable prompt generation and more robust cross-modal alignment.

\bibliographystyle{unsrtnat}
\bibliography{references}

\newpage
\appendix

\section{Details of point prompt generation}
\label{app:mpg_details}

The main paper describes the overall point-prompt generation process in MPG. Here, we provide the implementation details. Given the cross-modal similarity map $\mathbf{S}(\mathbf{x})$ produced by BiomedCLIP, where $\mathbf{x}$ denotes a spatial location, we first construct high-confidence candidate regions and then sample positive and negative prompt points from high-confidence foreground cores and neighboring hard-background regions, respectively.

MPG first computes a sample-adaptive threshold according to the similarity distribution. We use a quantile threshold with $q=0.90$ and constrain it to be non-negative:
\begin{equation}
    \mathbf{R}_{fg}(\mathbf{x})
    =
    \mathbb{I}
    \left[
    \mathbf{S}(\mathbf{x})
    \geq
    \max
    \left(
    Q_{0.90}(\mathbf{S}), 0
    \right)
    \right],
    \label{eq:app_point_high_region}
\end{equation}
where $Q_{0.90}(\mathbf{S})$ denotes the $0.90$ quantile of the similarity map and $\mathbb{I}[\cdot]$ is the indicator function.

We then perform connected-component analysis and remove isolated regions smaller than $16$ pixels. For each valid component, its area and peak similarity response are recorded, and at most two main candidate components are retained. This suppresses isolated false activations and provides stable spatial support for subsequent prompt sampling.

Before morphological processing, the candidate regions are represented at the $512\times512$ input resolution required by MedSAM2. Therefore, the morphological kernel sizes and point-distance constraints described below are defined in this spatial resolution.

Positive prompt points are sampled from high-confidence foreground cores. For each retained component $\mathbf{R}_{cc}$, morphological erosion is applied:
\begin{equation}
    \mathbf{R}_{core}
    =
    \operatorname{Erode}
    \left(
    \mathbf{R}_{cc}, k_e
    \right),
    \label{eq:app_point_core_region}
\end{equation}
where the erosion kernel size is $k_e=5$ with two iterations. If the eroded region contains fewer than $4$ pixels, the erosion strength is gradually reduced. If no valid foreground core remains, the original connected component is used as a fallback. Positive points are preferentially selected from high-response locations in $\mathbf{R}_{core}$. We first sample at most one point from each retained component and then supplement from the union of foreground cores until $N_+=2$ positive points are obtained.

Negative points are sampled from nearby hard-background regions. For each retained component, we construct a local background ring using dilation kernels of $17$ and $27$ pixels:
\begin{equation}
    \mathbf{R}_{ring}
    =
    \operatorname{Dilate}
    \left(
    \mathbf{R}_{cc}, 27
    \right)
    \setminus
    \left(
    \operatorname{Dilate}
    \left(
    \mathbf{R}_{cc}, 17
    \right)
    \cup
    \mathbf{R}_{sel}
    \right),
    \label{eq:app_point_negative_ring}
\end{equation}
where $\mathbf{R}_{sel}$ denotes the union of all retained foreground candidate components. If $\mathbf{R}_{ring}$ is empty, the inner dilated region excluding the retained foreground components is used as a fallback.

During sampling, at most $128$ candidate points are retained from each candidate region. Positive points are selected in descending order of similarity response, while negative points are preferentially sampled from high-response locations in the background ring to form hard negatives. A minimum distance of $12$ pixels is enforced to avoid excessive clustering. We use $N_+=2$ positive and $N_-=2$ negative points:
\begin{equation}
    \mathbf{P}_{pt}
    =
    \left\{
    \left(
    \mathbf{p}_i^+, 1
    \right)
    \right\}_{i=1}^{2}
    \cup
    \left\{
    \left(
    \mathbf{p}_j^-, 0
    \right)
    \right\}_{j=1}^{2},
    \label{eq:app_point_prompt_set}
\end{equation}
where $\mathbf{p}_i^+$ and $\mathbf{p}_j^-$ denote positive and negative points, and labels $1$ and $0$ indicate foreground and background, respectively.

\section{Loss function details}
\label{app:loss_details}

This section provides the detailed definitions of the Binary Cross-Entropy (BCE) and Dice losses used in RADAR. Let $\mathbf{Y}\in\{0,1\}^{H\times W}$ denote the ground-truth mask and $\mathbf{P}\in[0,1]^{H\times W}$ a predicted probability map. The spatial domain is denoted by $\Omega$.

\paragraph{Binary Cross-Entropy loss.}

The BCE loss is defined as
\begin{equation}
    \mathcal{L}_{BCE}(\mathbf{P},\mathbf{Y})
    =
    -
    \frac{1}{|\Omega|}
    \sum_{\mathbf{x}\in\Omega}
    \left[
    \mathbf{Y}(\mathbf{x})
    \log
    \left(
    \mathbf{P}(\mathbf{x})+\epsilon
    \right)
    +
    \left(
    1-\mathbf{Y}(\mathbf{x})
    \right)
    \log
    \left(
    1-\mathbf{P}(\mathbf{x})+\epsilon
    \right)
    \right],
    \label{eq:app_bce_loss}
\end{equation}
where $\epsilon$ is a small constant for numerical stability.

\paragraph{Dice loss.}

The Dice loss is defined as
\begin{equation}
    \mathcal{L}_{Dice}(\mathbf{P},\mathbf{Y})
    =
    1
    -
    \frac{
    2
    \sum_{\mathbf{x}\in\Omega}
    \mathbf{P}(\mathbf{x})\mathbf{Y}(\mathbf{x})
    +
    \epsilon
    }{
    \sum_{\mathbf{x}\in\Omega}
    \mathbf{P}(\mathbf{x})
    +
    \sum_{\mathbf{x}\in\Omega}
    \mathbf{Y}(\mathbf{x})
    +
    \epsilon
    }.
    \label{eq:app_dice_loss}
\end{equation}

The final MedSAM2 segmentation loss is
\begin{equation}
    \mathcal{L}_{seg}
    =
    0.5
    \mathcal{L}_{BCE}
    \left(
    \mathbf{Y}_{pred},\mathbf{Y}
    \right)
    +
    0.5
    \mathcal{L}_{Dice}
    \left(
    \mathbf{Y}_{pred},\mathbf{Y}
    \right).
    \label{eq:app_seg_loss}
\end{equation}

For similarity-map supervision, the cross-modal cosine similarity is converted into a probability map using a sigmoid function:
\begin{equation}
    \mathbf{P}_{sim}(\mathbf{x})
    =
    \sigma
    \left(
    \mathbf{S}(\mathbf{x})
    \right),
    \label{eq:app_similarity_probability}
\end{equation}
where $\sigma(\cdot)$ denotes the sigmoid function. The ground-truth mask is resized to the corresponding spatial resolution, denoted as $\mathbf{Y}_{sim}$, and the auxiliary similarity loss follows the same BCE-Dice formulation:
\begin{equation}
    \mathcal{L}_{sim}
    =
    0.5
    \mathcal{L}_{BCE}
    \left(
    \mathbf{P}_{sim},\mathbf{Y}_{sim}
    \right)
    +
    0.5
    \mathcal{L}_{Dice}
    \left(
    \mathbf{P}_{sim},\mathbf{Y}_{sim}
    \right).
    \label{eq:app_similarity_loss}
\end{equation}

Together with the structured dense representation alignment loss $\mathcal{L}_{align}$, the total objective is
\begin{equation}
    \mathcal{L}_{total}
    =
    \mathcal{L}_{seg}
    +
    \lambda_{sim}\mathcal{L}_{sim}
    +
    \lambda_{align}\mathcal{L}_{align}.
    \label{eq:app_total_loss}
\end{equation}
We set $\lambda_{sim}=0.2$ and $\lambda_{align}=0.2$ in all experiments.

\section{Dataset and evaluation metrics}
\label{app:datasets_metrics}

\subsection{Dataset details}
\label{app:datasets}

Our main evaluation uses three public medical image segmentation datasets: QaTa-COV19, MosMedData+, and Kvasir.

QaTa-COV19~\citep{Degerli2022Osegnet} is a chest X-ray infection segmentation dataset for COVID-19-related lesions. We follow the data split and text annotations provided by~\citep{li2024LViT}. The training, validation, and testing sets contain 5716, 1429, and 2113 images, respectively.

MosMedData+~\citep{morozov2020mosmeddata,hofmanninger2020automatic} is a chest CT infection segmentation dataset with pixel-level infection annotations. We follow the split and text annotation protocol of~\citep{li2024LViT}. The training, validation, and testing sets contain 2183, 273, and 273 CT slices, respectively.

Kvasir~\citep{Jha2020Kvasir} is a colonoscopy polyp segmentation dataset containing 1000 images. We follow the text annotations and experimental setting of~\citep{Zha_MAdapter_MICCAI2024}. The standard testing set is kept fixed, while the remaining images are divided into training and validation subsets with a 9:1 ratio using random seed 42, resulting in 810, 90, and 100 images for training, validation, and testing, respectively.

All compared methods are retrained and evaluated using the same data splits, preprocessing pipeline, and evaluation protocol. Text-guided methods additionally use the same image-text annotations. The dataset split statistics are summarized in Table~\ref{tab:app_dataset_split}.

\begin{table}[htbp]
\caption{Statistics of the dataset splits used for training, validation, and testing.}
\label{tab:app_dataset_split}
\centering
\small
\setlength{\tabcolsep}{6pt}
\renewcommand{\arraystretch}{1.0}
\resizebox{\linewidth}{!}{
\begin{tabular}{lll|ccc|c}
\toprule
Dataset & Modality & Task & Training & Validation & Testing & Total \\
\midrule
QaTa-COV19~\citep{Degerli2022Osegnet}
& X-ray & Chest infection & 5716 & 1429 & 2113 & 9258 \\
MosMedData+~\citep{morozov2020mosmeddata,hofmanninger2020automatic}
& CT & Chest infection & 2183 & 273 & 273 & 2729 \\
Kvasir~\citep{Jha2020Kvasir}
& Colonoscopy & Polyp segmentation & 810 & 90 & 100 & 1000 \\
\bottomrule
\end{tabular}
}
\end{table}

\subsection{Evaluation metrics}
\label{app:metrics}

We use Dice, mean Intersection over Union (mIoU), 95th percentile Hausdorff Distance (HD95), and Average Symmetric Surface Distance (ASSD). Predicted probability maps are binarized at a threshold of $0.5$ before evaluation. All metrics are computed per test sample and then averaged over the test set. Dice and mIoU are reported as percentages, while HD95 and ASSD are measured in pixels.

Let $P_i$ and $G_i$ denote the predicted and ground-truth foreground pixel sets of the $i$-th test sample, respectively. Dice is defined as
\begin{equation}
    \mathrm{Dice}
    =
    \frac{1}{N}
    \sum_{i=1}^{N}
    \frac{
    2|P_i\cap G_i|+\epsilon
    }{
    |P_i|+|G_i|+\epsilon
    },
    \label{eq:app_dice_metric}
\end{equation}
where $N$ denotes the number of test samples.

In our binary segmentation setting, mIoU denotes the sample-averaged foreground IoU:
\begin{equation}
    \mathrm{mIoU}
    =
    \frac{1}{N}
    \sum_{i=1}^{N}
    \frac{
    |P_i\cap G_i|+\epsilon
    }{
    |P_i\cup G_i|+\epsilon
    }.
    \label{eq:app_miou_metric}
\end{equation}

For boundary metrics, let $\partial P_i$ and $\partial G_i$ denote the predicted and ground-truth boundary point sets. The shortest Euclidean distance from a point $a$ to a boundary set $B$ is
\begin{equation}
    d(a,B)
    =
    \min_{b\in B}
    \|a-b\|_2.
    \label{eq:app_boundary_distance}
\end{equation}

For samples with non-empty predicted and ground-truth boundaries, the bidirectional distance set is
\begin{equation}
    D_i
    =
    \{d(a,\partial G_i)\mid a\in\partial P_i\}
    \cup
    \{d(b,\partial P_i)\mid b\in\partial G_i\}.
    \label{eq:app_bidirectional_distance}
\end{equation}

Since distance metrics are undefined when either boundary is empty, such samples are excluded from HD95 and ASSD:
\begin{equation}
    \mathcal{V}
    =
    \left\{
    i
    \mid
    \partial P_i \neq \emptyset,\,
    \partial G_i \neq \emptyset
    \right\}.
    \label{eq:app_valid_distance_set}
\end{equation}
The same exclusion rule is applied consistently to all compared methods.

HD95 is defined as
\begin{equation}
    \mathrm{HD95}
    =
    \frac{1}{|\mathcal{V}|}
    \sum_{i\in\mathcal{V}}
    \mathrm{Percentile}_{95}(D_i),
    \label{eq:app_hd95_metric}
\end{equation}
and ASSD as
\begin{equation}
    \mathrm{ASSD}
    =
    \frac{1}{|\mathcal{V}|}
    \sum_{i\in\mathcal{V}}
    \frac{
    \sum_{a\in\partial P_i} d(a,\partial G_i)
    +
    \sum_{b\in\partial G_i} d(b,\partial P_i)
    }{
    |\partial P_i|+|\partial G_i|
    }.
    \label{eq:app_assd_metric}
\end{equation}

Higher Dice and mIoU indicate better region overlap, while lower HD95 and ASSD indicate smaller boundary errors.

\section{Additional analysis}
\label{app:additional_analysis}

\subsection{Robustness to text perturbations}
\label{app:text_robustness}

We evaluate the dependence of RADAR on textual guidance by fixing the trained checkpoint and all test images while changing only the test-time text. Incomplete prompts remove disambiguating location, number, or extent attributes and therefore represent under-specified descriptions.

\begin{table}[htbp]
\caption{
Robustness to different test-time text conditions on QaTa-COV19 and MosMedData+. The model checkpoint and input images are fixed, and only the textual descriptions are changed. Best and second-best results are \textbf{bolded} and \underline{underlined}, respectively.
}
\label{tab:app_text_robustness}
\centering
\scriptsize
\setlength{\tabcolsep}{2.5pt}
\renewcommand{\arraystretch}{1.0}
\resizebox{0.85\linewidth}{!}{
\begin{tabular}{l|cccc|cccc}
\toprule
\multirow{2}{*}{Text type}
& \multicolumn{4}{c|}{QaTa-COV19}
& \multicolumn{4}{c}{MosMedData+} \\
\cmidrule(lr){2-5}
\cmidrule(lr){6-9}
& Dice $\uparrow$ & mIoU $\uparrow$ & HD95 $\downarrow$ & ASSD $\downarrow$
& Dice $\uparrow$ & mIoU $\uparrow$ & HD95 $\downarrow$ & ASSD $\downarrow$ \\
\midrule

Original
& \textbf{84.12} & \textbf{75.32} & \textbf{14.27} & \textbf{4.09}
& \textbf{75.47} & \textbf{62.43} & \textbf{13.16} & \textbf{3.26} \\

Empty
& 76.90 & 66.92 & 33.90 & 10.13
& 67.80 & 54.40 & 20.49 & 5.23 \\

Incomplete
& 82.01 & 71.95 & 20.32 & 5.67
& 73.72 & 60.53 & 13.77 & 3.60 \\

Mild Noise
& \underline{83.57} & \underline{74.72} & \underline{17.54} & \underline{4.85}
& \underline{75.02} & \underline{61.97} & \underline{13.40} & \underline{3.28} \\

Synonym
& 83.26 & 74.32 & 19.95 & 5.39
& 74.56 & 61.41 & 13.79 & 3.33 \\

Wrong
& 70.33 & 58.38 & 46.25 & 13.98
& 70.63 & 57.00 & 26.34 & 6.86 \\

\bottomrule
\end{tabular}
}
\end{table}

As shown in Table~\ref{tab:app_text_robustness}, mild noise and synonym substitutions cause relatively small performance drops, whereas incomplete descriptions produce larger degradation. Empty text reduces Dice by 7.22 and 7.67 percentage points on QaTa-COV19 and MosMedData+, respectively, with substantially increased boundary errors. Incorrect text further degrades regional and boundary quality. These results show that RADAR is reasonably robust to semantic-preserving text variations but still depends on correct image-text correspondence.

\subsection{Comparison with learnable prompt generation}
\label{app:learnable_mpg}

To examine whether the handcrafted spatial prompt conversion in MPG can be replaced by alternative strategies, we keep BiomedCLIP, MedSAM2, GRDA, SDRA, losses, and all other training settings fixed. RADAR-SimMask directly uses the normalized similarity response as the mask prompt, while RADAR-Extrema samples extrema from the response map as spatial prompts. RADAR-LMPG instead concatenates the final BiomedCLIP spatial features with the similarity map and uses a three-layer convolutional head to predict a mask prompt and positive/negative heatmaps. Spatial softmax and differentiable soft-argmax are then used to obtain point coordinates.

As shown in Table~\ref{tab:app_learnable_mpg}, direct similarity-map and extrema prompts are more sensitive to false activations and uncertain boundaries. RADAR-LMPG is a viable learnable alternative and slightly improves Kvasir Dice and mIoU, but performs worse on all MosMedData+ metrics and on Kvasir boundary metrics while increasing trainable parameters from 0.74M to 0.91M (+22.97\%). The original MPG therefore provides a more balanced performance--parameter trade-off in the current setting.

\begin{table}[htbp]
\caption{
Comparison of different prompt-generation strategies on MosMedData+ and Kvasir. Best and second-best results are \textbf{bolded} and \underline{underlined}, respectively.
}
\label{tab:app_learnable_mpg}
\centering
\scriptsize
\setlength{\tabcolsep}{2.2pt}
\renewcommand{\arraystretch}{1.0}
\resizebox{\linewidth}{!}{
\begin{tabular}{lc|cccc|cccc}
\toprule
\multirow{2}{*}{Method}
& \multirow{2}{*}{Trainable Params}
& \multicolumn{4}{c|}{MosMedData+}
& \multicolumn{4}{c}{Kvasir} \\
\cmidrule(lr){3-6}
\cmidrule(lr){7-10}
& & Dice $\uparrow$ & mIoU $\uparrow$ & HD95 $\downarrow$ & ASSD $\downarrow$
& Dice $\uparrow$ & mIoU $\uparrow$ & HD95 $\downarrow$ & ASSD $\downarrow$ \\
\midrule

RADAR-SimMask
& 0.74M
& 74.81 & 61.68 & 14.05 & 3.33
& 91.00 & 84.98 & 19.49 & 4.72 \\

RADAR-Extrema
& 0.74M
& \underline{75.19} & \underline{61.98} & 13.71 & 3.57
& 91.55 & 85.75 & 16.84 & 4.31 \\

RADAR-LMPG
& 0.91M
& 75.10 & 61.97 & \underline{13.38} & \underline{3.29}
& \textbf{92.24} & \textbf{86.74} & \underline{16.73} & \underline{4.03} \\

RADAR
& 0.74M
& \textbf{75.47} & \textbf{62.43} & \textbf{13.16} & \textbf{3.26}
& \underline{92.15} & \underline{86.65} & \textbf{14.79} & \textbf{3.91} \\

\bottomrule
\end{tabular}
}
\end{table}

\section{Additional ablation studies}
\label{app:additional_ablation}

\subsection{Ablation study on LoRA injection layers}
\label{app:lora_ablation}

Table~\ref{tab:lora_layers_ablation} compares different LoRA injection ranges. Full-layer adaptation slightly improves Kvasir overlap but increases trainable parameters to 1.23M, while overly narrow injection ranges reduce cross-dataset stability. The adopted configuration applies LoRA to all MedSAM2 image-encoder layers and the last four BiomedCLIP visual/text layers, providing a favorable accuracy--parameter trade-off with 0.74M trainable parameters.

\begin{table}[htbp]
\caption{
Ablation study on LoRA injection layers. S-I, B-V, and B-T denote the MedSAM2 image encoder, BiomedCLIP visual encoder, and BiomedCLIP text encoder, respectively. The MedSAM2 decoder LoRA layers are fixed to S-D 0--1 in all settings. Best and second-best results are \textbf{bolded} and \underline{underlined}, respectively.
}
\label{tab:lora_layers_ablation}
\centering
\scriptsize
\setlength{\tabcolsep}{2.0pt}
\renewcommand{\arraystretch}{1.0}
\resizebox{\linewidth}{!}{
\begin{tabular}{llc|cccc|cccc}
\toprule
\multirow{2}{*}{Setting}
& \multirow{2}{*}{LoRA injection layers}
& \multirow{2}{*}{Trainable Params $\downarrow$}
& \multicolumn{4}{c|}{MosMedData+}
& \multicolumn{4}{c}{Kvasir} \\
\cmidrule(lr){4-7}
\cmidrule(lr){8-11}
& &
& Dice $\uparrow$ & mIoU $\uparrow$ & HD95 $\downarrow$ & ASSD $\downarrow$
& Dice $\uparrow$ & mIoU $\uparrow$ & HD95 $\downarrow$ & ASSD $\downarrow$ \\
\midrule

All layers
& S-I 0--11, B-V/B-T 0--11
& 1.23M
& 74.67 & 61.44 & 14.29 & 3.49
& \textbf{92.29} & \textbf{86.92} & \underline{14.92} & \textbf{3.76} \\

BioCLIP late-6
& S-I 0--11, B-V/B-T 6--11
& 0.86M
& 73.52 & 60.19 & 14.35 & 3.88
& 92.03 & 86.63 & 15.17 & \underline{3.90} \\

BioCLIP last-2
& S-I 0--11, B-V/B-T 10--11
& \textbf{0.62M}
& \underline{74.78} & \underline{61.51} & 13.77 & 3.53
& 91.52 & 85.62 & 17.45 & 4.56 \\

MedSAM2 late-6
& S-I 6--11, B-V/B-T 0--11
& 0.69M
& 74.22 & 60.94 & \textbf{12.97} & \underline{3.28}
& 91.98 & 86.48 & 16.74 & 4.32 \\

MedSAM2 late-4
& S-I 8--11, B-V/B-T 0--11
& \underline{0.67M}
& 72.17 & 58.43 & 13.20 & 3.63
& 91.89 & \underline{86.86} & 16.88 & 4.45 \\

\midrule

RADAR (Ours)
& S-I 0--11, B-V/B-T 8--11
& 0.74M
& \textbf{75.47} & \textbf{62.43} & \underline{13.16} & \textbf{3.26}
& \underline{92.15} & 86.65 & \textbf{14.79} & 3.91 \\

\bottomrule
\end{tabular}
}
\end{table}

\section{Extended visualization analysis of SDRA}
\label{app:sdra_visualization}

\begin{figure}[htbp]
  \centering
  \includegraphics[width=0.9\linewidth]{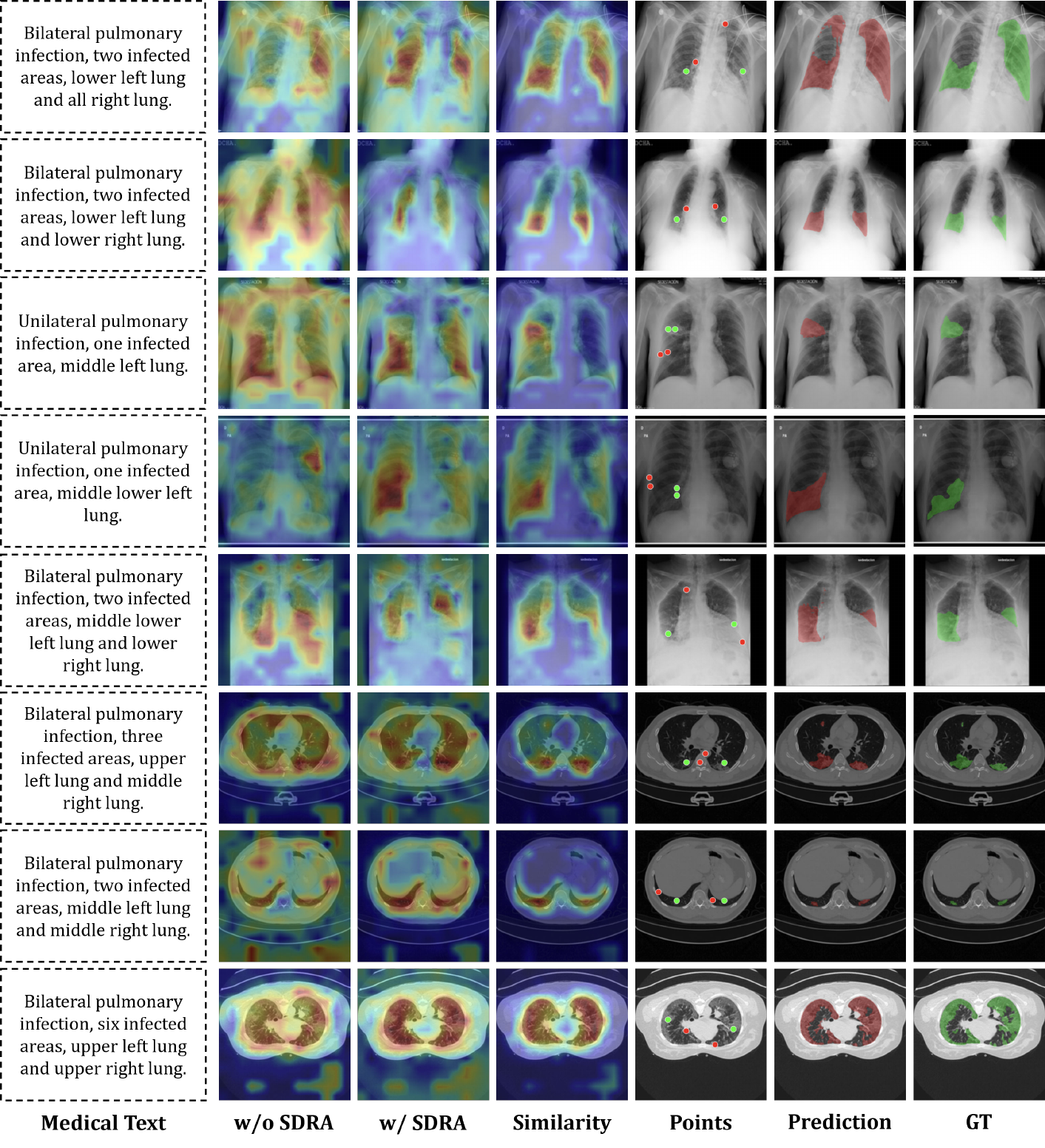}
  \caption{
  Extended visualization of SDRA on QaTa-COV19 and MosMedData+. The columns show the medical text, BiomedCLIP dense features without and with SDRA, text-image similarity maps, generated point prompts, predictions, and ground-truth masks.
  }
  \label{fig:workflow_visualization}
\end{figure}

Figure~\ref{fig:workflow_visualization} further illustrates the effect of SDRA on BiomedCLIP dense representations and subsequent prompt generation using representative samples from QaTa-COV19 and MosMedData+. The visualization includes dense features before and after SDRA, text-image similarity maps, generated point prompts, final predictions, and ground-truth masks.

Without SDRA, high-response regions are often distributed across non-target lung regions, bone structures, and visually similar backgrounds. After introducing SDRA, responses become more concentrated around lesion regions with fewer irrelevant activations. The resulting similarity maps and point prompts also show better spatial consistency with the target regions. These observations suggest that SDRA improves the spatial concentration of cross-modal responses and reduces noisy prompt propagation, providing more informative spatial cues for MedSAM2 decoding.

\end{document}